\documentclass{article}

\usepackage[final]{neurips_2022}

\usepackage[utf8]{inputenc} % allow utf-8 input
\usepackage[T1]{fontenc}    % use 8-bit T1 fonts
\usepackage{hyperref}       % hyperlinks
\usepackage{url}            % simple URL typesetting
\usepackage{booktabs}       % professional-quality tables
\usepackage{amsfonts}       % blackboard math symbols
\usepackage{nicefrac}       % compact symbols for 1/2, etc.
\usepackage{microtype}      % microtypography
\usepackage{xcolor}         % colors

\usepackage{algorithm}
\usepackage{algorithmic}
\usepackage{enumitem}
\usepackage{amsmath,amssymb} % define this before the line numbering.
\usepackage{xspace}
\usepackage{graphicx}
\usepackage{bbm}
\usepackage{multirow}
\usepackage{subcaption}
\usepackage{wrapfig}
\usepackage{bbm}
\definecolor{citecolor}{HTML}{0071bc}
\hypersetup{
  colorlinks,
  citecolor=citecolor,
  linkcolor=red
}

\newlength\savewidth\newcommand\shline{\noalign{\global\savewidth\arrayrulewidth
  \global\arrayrulewidth 1pt}\hline\noalign{\global\arrayrulewidth\savewidth}}

\newcommand{\norm}[1]{\left\lVert#1\right\rVert}

\newcommand{\revise}[1]{\textcolor{black}{#1}}
\usepackage[normalem]{ulem}
\useunder{\uline}{\ul}{}

\def\eg{\emph{e.g.}}
\def\ie{\emph{i.e.}}

\def\etc{\emph{etc.}} 
\def\mD{{\mathcal D}}

\def\mO{{\mathcal O}}

\def\mQ{{\mathcal Q}}

\def\mS{{\mathcal S}}

\def\mU{{\mathcal U}}

\def\0{{\bf 0}}
\def\1{{\bf 1}}

\def\bA{{\bf A}}

\def\bQ{{\bf Q}}

\def\bW{{\bf W}}
\def\bX{{\bf X}}
\def\bY{{\bf Y}}

\def\ba{{\bf a}}

\def\bg{{\bf g}}

\def\bk{{\bf k}}

\def\bq{{\bf q}}

\def\bu{{\bf u}}
\def\bv{{\bf v}}

\def\bx{{\bf x}}
\def\by{{\bf y}}

\def\citep{\cite}
\def\citet{\cite}

\newcommand{\methodshortname}{EcoFormer\xspace}

\title{EcoFormer: Energy-Saving Attention \\ with Linear Complexity}

\author{%
Jing Liu\thanks{Authors contributed equally.}
\quad Zizheng Pan$^{*}$
\quad Haoyu He 
\quad Jianfei Cai 
\quad Bohan Zhuang\thanks{Corresponding author. Email: $\tt bohan.zhuang@monash.edu$} \\[0.2cm]
  Department of Data Science \& AI, Monash University, Australia
}

\begin{document}

\maketitle

\begin{abstract}
Transformer is a transformative framework for deep learning which models sequential data and has achieved remarkable performance on a wide range of tasks, but with high computational and energy cost. To improve its efficiency, a popular choice is to compress the models via binarization which constrains the floating-point values into binary ones to save resource consumption owing to cheap bitwise operations  significantly. However, existing binarization methods only aim at minimizing the information loss for the input distribution statistically, while ignoring the pairwise similarity modeling at the core of the attention mechanism. To this end, we propose a new binarization paradigm customized to high-dimensional softmax attention via kernelized hashing, called EcoFormer, to map the original queries and keys into low-dimensional binary codes in Hamming space. The kernelized hash functions are learned to match the ground-truth similarity relations extracted from the attention map in a self-supervised way. Based on the equivalence between the inner product of binary codes and the Hamming distance as well as the associative property of matrix multiplication, we can approximate the attention in linear complexity by expressing it as a dot-product of binary codes. Moreover, the compact binary representations of queries and keys in EcoFormer enable us to replace most of the expensive multiply-accumulate operations in attention with simple accumulations to save considerable on-chip energy footprint on edge devices. Extensive experiments on both vision and language tasks show that EcoFormer consistently achieves comparable performance with standard attentions while consuming much fewer resources. For example, based on PVTv2-B0 and ImageNet-1K, EcoFormer achieves a 73\% reduction in on-chip energy footprint with only a slight performance drop of 0.33\% compared to the standard attention. Code is available at \url{https://github.com/ziplab/EcoFormer}.
\end{abstract}

\section{Introduction}
Recently, Transformers~\cite{vaswani2017attention} have shown rapid and exciting progress in natural language processing (NLP)~\cite{devlin2019bert,dehghani2018universal} and computer vision (CV)~\cite{dosovitskiy2021an,touvron2021training} due to its extraordinary representational power. Compared with convolutional neural networks (CNNs)~\cite{krizhevsky2012imagenet}, Transformer models are generally more scalable to massive amounts of data and better at capturing long-dependency global information with less inductive bias, thus achieving better performance in many tasks~\cite{jaegle2021perceiver,liu2021swin}. 
However, the efficiency bottlenecks, especially the high energy consumption, greatly hamper the massive deployment of Transformer models to resource-constrained edge devices, such as mobile phones and unmanned aerial vehicles, for solving a variety of real-world applications.
\begin{figure}[t!]
	\centering
	\includegraphics[width=0.95\linewidth]{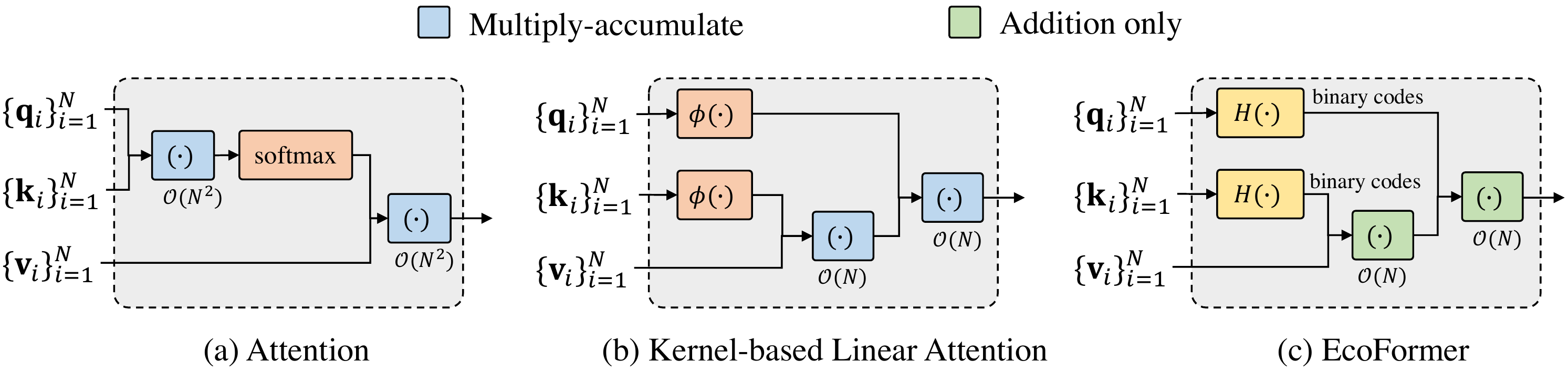}
	\caption{Computational graphs for standard attention (left), kernel-based linear attention (middle) and the proposed EcoFormer based on kernelized hashing (right).}
	\label{fig:framework}
	\vspace{-6mm}
\end{figure}

\begin{table}[!tb]
\caption{Energy cost for different operations (on 45nm CMOS technology)~\cite{han2016eie,horowitz20141,lian2019high}.
}
\centering
\renewcommand\arraystretch{1.2}
\scalebox{0.82}
{
\begin{tabular}{c|cccc}
Operation & 16-bit FP Add & 16-bit FP Mult & 32-bit FP Add & 32-bit FP Mult \\
\shline
Energy (pJ) & 0.4 & 1.1 & 0.9 & 3.7 \\
Area ($\mu m^2$) & 1,360 & 1,640 & 4,184 & 7,700 \\
\end{tabular}
}
\label{table:energy_area_cost}
 \vspace{-7mm}
\end{table}

To reduce the energy consumption, quantization has been actively studied to lower the bit-width representation of network weights~\cite{courbariaux2015binaryconnect,shen2020q,zadeh2020gobo} and/or activations~\cite{hubara2017quantized,zhou2016dorefa,zafrir2019q8bert}. With the most aggressive bit-width, binary quantization~\cite{rastegari2016xnor,bai2021binarybert,qin2022bibert} has attracted much attention since it enables efficient bit-wise operations by representing values with a single bit (\eg, +1 or -1). When we only binarize weights in analogous to BinaryConnect \cite{courbariaux2015binaryconnect}, as shown in Table~\ref{table:energy_area_cost}, it brings great benefits to dedicated hardware by replacing a large number of energy-hungry multiply-accumulate operations with simple energy-efficient accumulations, which
saves significant on-chip area and energy required to run inference with Transformers, making them feasible to be deployed on mobile platforms with limited resources. However, the conventional binarization process typically targets at minimizing quantization errors between the original full-precision data distribution and the quantized Bernoulli distribution statistically. In other words, each token is binarized separately, where the binary representations may not well preserve the original similarity relations among tokens.
This motivates us to customize the binarization process to softmax attention, the core mechanism in Transformer that encodes the pairwise similarity between tokens.
To this end, we can adapt the well-established hashing methods, to map the high-dimensional queries and keys into compact binary codes (\eg, 16-bits) that are able to preserve the similarity relations in Hamming space.
A simple solution is to use the locality-sensitive hashing (LSH)~\cite{andoni2015practical}  to substitute the binary quantization counterparts. 

Nevertheless, another energy bottleneck exists in Transformers. Specially, given a sequence of tokens, the softmax attention obtains the attention weights by computing the inner product between a query token and all key tokens, leading to the quadratic time complexity $\mO\left(N^{2}\right)$ regarding the number of tokens $N$, as shown in Figure~\ref{fig:framework} (a). This problem is even worse for a long sequence length $N$, especially for high resolution images in dense prediction tasks. To reduce the complexity of the softmax attention, some prior works propose to express the attention as a linear dot product of kernelized feature embeddings~\cite{peng2020random, choromanski2021rethinking}. With the associative property of matrix multiplication, the attention operation can be approximated in linear complexity $\mO(N)$, as illustrated by Figure \ref{fig:framework} (b).

Based on the hashing mechanism and kernel-based formulation of attention, we devise a simple yet effective energy-saving attention, called \methodshortname, which is shown in Figure~\ref{fig:framework} (c).
In particular, we propose to use kernelized hashing with RBF kernel to map the queries and keys to compact binary codes. 
The resulting codes are valid for similarity preserving based on the good property that the codes' inner product (\ie, Hamming affinity) and Hamming distance have one-to-one correspondence \cite{liu2012supervised}. Thanks to the associative property of the linear dot-product between the binary codes, the kernelized hashing attention is in \emph{linear complexity} with significant energy saving.
Moreover, the pairwise similarity matrix in attention can be directly used to obtain the supervision labels for hash function learning, delivering a novel \emph{self-supervised} hashing paradigm. By maximizing the Hamming affinity on the similar pairs of tokens and simultaneously minimizing on the dissimilar pairs of tokens, the pairwise similarity relations between tokens can be preserved. With \emph{low-dimensional binary} queries and keys, we can replace most of the energy-hungry floating-point multiplications in attention with simple additions, which greatly saves the on-chip energy footprint.

To sum up, we make three main contributions: 1) We propose a new binarization paradigm to better preserve the pairwise similarity in softmax attention. In particular, we present EcoFormer, an energy-saving attention with linear complexity powered by kernelized hashing to map the queries and keys into compact binary codes. 2) We learn the kernelized hash functions based on the ground-truth Hamming affinity extracted from the attention scores in a self-supervised way. 
3) Extensive experiments on CIFAR-100, ImageNet-1K and Long Range Arena show that \methodshortname is able to significantly reduce the \revise{on-chip} energy cost while preserving the accuracy. 
For example, based on PVTv2-B0 and ImageNet-1K, \methodshortname achieves a 73\% reduction in \revise{on-chip} energy footprint with only a marginal performance drop of 0.33\% compared to the standard attention.

\vspace{-2mm}
\section{Related Work}
\noindent\textbf{Efficient attention mechanisms.}
To alleviate the quadratic computational cost for vanilla attention with respect to the number of tokens, much work has endeavored developing efficient attentions. One line of research performs attention only on the part of the tokens~\cite{liu2021swin, wang2021pyramid, beltagy2020longformer, reformer, linformer, daras2020smyrf, vyas2020fast, sun2021sparse}.
For instance, Reformer~\cite{reformer}, SMYRF~\cite{daras2020smyrf}, Fast Transformers~\cite{vyas2020fast} and LHA~\cite{sun2021sparse} restrict the attention to the most similar token pairs via hashing and reduces the computational complexity to $\mO(N \log N)$. 
Linformer~\cite{linformer} approximates the attention with low-rank factorization that reduces the length of the key and value. However, the computational complexity is dependent on the design for reducing tokens.
Another line of research speeds up the vanilla attention with kernel-based methods~\cite{choromanski2021rethinking, xiong2021nystromformer,lu2021soft, katharopoulos2020transformers, zhen2022cosformer}. For example, Performer~\cite{choromanski2021rethinking} approximates the softmax operation with orthogonal random features.  Nyströmformer~\cite{xiong2021nystromformer} and SOFT~\cite{lu2021soft} approximate the full self-attention matrix via matrix decomposition. 
Although impressive achievements have been achieved, how to develop attention that is highly energy-efficient remains under-explored, as multiplications dominate the \revise{on-chip} energy consumption. AdderNet variants~\cite{chen2020addernet, shu2021adder} replace the energy-hungry multiplication-based similarity measurement with the energy-efficient addition-based L1 distance and argue that additions can also provide powerful feature representations. Nevertheless, this heuristic approach brings a drastic performance drop. In contrast, we are from the perspective of binarization and propose to learn kernel-based hash functions using attention scores to map the original features into compact similarity-preserving binary codes in Hamming distance, which is energy-efficient and in linear complexity $\mO(N)$. With the low-dimensional binary queries and keys, our EcoFormer is able to replace most of the multiplications with simple accumulations.

\noindent\textbf{Hashing.}
Hashing is an efficient nearest neighbor search method by embedding the high-dimensional data into a similarity preserving low-dimensional binary codes, based on the intuition that highly similar data should be assigned the same hash key. 
Hashing methods can be roughly categorized into data-independent and data-dependent schemes. The former focuses on building random hash functions and locally sensitive hashing (LSH) \cite{gionis1999similarity, charikar2002similarity} is arguably the most representative one, which guarantees the sub-linear time similarity search and is followed by non-linear extensions such as hashing with kernels \cite{kulis2011kernelized} or on manifolds \cite{weiss2008spectral}. The latter can be further divided into unsupervised \cite{liu2011hashing, gong2012iterative, liu2014discrete}, semi-supervised \cite{norouzi2011minimal, wang2012semi} and supervised hashing \cite{liu2012supervised, zhuang2016fast, yan2020deep}.
When it comes to Transformers, as self-attention encodes the pairwise similarity among tokens, hashing is thus a natural choice to efficiently retrieve similar keys given a query. Reformer is such a pioneering work, which proposes to group similar tokens in a single hash bucket to form sparse self-attention. Our EcoFormer is fundamentally different from Reformer in three aspects: 1) Reformer relies on local attention lookups to reduce the complexity while our EcoFormer is designed from a numerical perspective, where the low-dimensional binary codes are used to save the multiplications; 2) Reformer is built upon LSH with linear mapping, which cannot deal with the 
kernel-based formulation of attention to scale linearly with the sequence length; 
3) Our hash functions are self-supervised by the pairwise affinity labels in attention, which are optimized in conjunction with network parameters and more accurate than unsupervised random projections. 

\noindent\textbf{Binary quantization.}
Binarization, an extreme quantization scheme, seeks to represent the vectors by binary codes. As a result, the computationally heavy matrix multiplications become light-weight bitwise operations (\ie, $\mathrm{xnor}$ and $\mathrm{popcount}$), yielding promising memory saving and acceleration. In general, to make binary neural networks \cite{hubara2016binarized} reliable in accuracy, current research targets at tackling two main challenges. The first challenge is to minimize the quantization error, basically based on learning the scaling factors \cite{rastegari2016xnor, bulat2019xnor}, parameterizing the quantization range and/or intervals \cite{jung2019learning, esser2020learned}, and ensembling multiple binary bases~\cite{lin2017towards, zhuang2019structured}, \etc~ Another category of studies focus on solving the non-differentiable optimization problem due to the discretization process via training with regularization \cite{ding2019regularizing}, knowledge distillation \cite{mishra2017apprentice, qin2022bibert}, relaxed optimization \cite{hou2018loss, bai2019proxquant}, appending full-precision branches \cite{Liu_2018_ECCV, martinez2020training} and so on. Apart from CNNs, there are some recent pioneering attempts targeting on binarizing Transformers. For example, BinaryBERT~\cite{bai2021binarybert} proposes to push Transformer quantization to the limit by weight binarization.
BiBERT~\cite{qin2022bibert} quantizes both weights, embeddings and activations of BERT~\cite{devlin2019bert} to 1-bit and achieves considerable savings on FLOPs and model size, but still has obvious performance drop.
In contrast, we propose to customize the binarization paradigm to softmax attention from the hashing perspective, preserving high-fidelity pairwise similarity information in compact binary codes which are used to deliver linear-complexity, energy-efficient self-attention.

\vspace{-2mm}
\section{Preliminaries}
\subsection{Attention Mechanism}
\label{sec:attention}
Let $\bX \in \mathbb{R}^{N \times D}$ be the input sequence into a standard multi-head self-attention (MSA) layer, where $N$ is the length of the input sequence and $D$ is the number of hidden dimensions. A standard MSA layer calculates a sequence of query, key and value vectors with three learnable projection matrices $\bW_q, \bW_k, \bW_v \in \mathbb{R}^{D \times D_p}$, which can be formulated as
\begin{equation} \label{eq:qkv_proj}
{\begin{array}{ll}
    \{\bq_t\}^N_{t=1} = \bX\bW_q,
    \{\bk_t\}^N_{t=1} = \bX\bW_k,
    \{\bv_t\}^N_{t=1} = \bX\bW_v,
\end{array}}
\end{equation}
where $D_p$ refers to the number of dimensions for each head. For each query vector, the attention output is a weighted-sum over all value vectors as
\begin{equation} \label{eq:scale_product}
    \mathrm{Attention}(\bq_t, \{\bk_i\}, \{\bv_i\}) = \sum_{i}\frac{\mathrm{exp}(\bq_t \cdot \bk_i / \tau)}{\sum_{j}{\mathrm{exp}(\bq_t \cdot \bk_j / \tau)}} \bv_i,
\end{equation}
where $\tau$ is the temperature for controlling the flatness of softmax and $\exp(\langle\boldsymbol{\cdot}\,,\boldsymbol{\cdot}\rangle)$ is an exponential function. With $N$ token, the computation of attention has a quadratic complexity of $O(N^2)$ in both space and time, which results in huge computational cost when dealing with long sequences.

\subsection{Kernel-based Linear Attention}
\label{sec:linear_attention}
The idea behind kernel-based linear attention is to express the similarity measure in Eq.~(\ref{eq:scale_product}) as a linear dot-product of kernel embeddings, such as polynomial kernel, exponential or RBF kernel. A particular choice is to employ the finite random mapping~\cite{NIPS2007_013a006f} $\boldsymbol{\phi}(\cdot)$ to approximate the infinite dimensional RBF kernel.
Then, according to the theorem from Rahimi~\cite{NIPS2007_013a006f}, the inner product between a pair of transformed vectors $\bx$ and $\by$ with $\boldsymbol{\phi}(\cdot)$ can approximate a Guassian RBF kernel.
This gives rise to an unbiased estimation to $\exp(\langle\boldsymbol{\cdot}\,,\boldsymbol{\cdot}\rangle)$ in Eq.~(\ref{eq:scale_product}), which can be expressed as 
\begin{equation}\label{eq:approx_dot_exp}
\begin{split}
{\begin{array}{ll}
    \exp\left({\bx\cdot\by / \sigma^2} \right)
    &= \exp \left({\norm{\bx}^2/2\sigma^2 + \norm{\by}^2/2\sigma^2}\right) \exp\left({-\norm{\bx-\by}^2 / 2\sigma^2} \right) \\
    &\approx \exp \left({\norm{\bx}^2/2\sigma^2 + \norm{\by}^2/2\sigma^2}\right)\ 
    \boldsymbol{\phi}\left(\bx\right)^\top \boldsymbol{\phi}\left(\by\right).
\end{array}}
\end{split}
\end{equation}
Assume that the queries and keys are unit vectors, then the attention computation in Eq.~(\ref{eq:scale_product}) can be approximated by 
\begin{subequations}
\label{eq:approx_softmax}
\begin{align}
        \mathrm{Attention}(\bq_t, \{\bk_i\}, \{\bv_i\}) 
        % =  \\
        &\approx     \sum_i\frac{\boldsymbol{\phi}\left(\bq_t\right)^\top \boldsymbol{\phi}\left(\bk_i\right)\bv_i }
        {\sum_j\boldsymbol{\phi}\left(\bq_t\right)^\top \boldsymbol{\phi}\left(\bk_j\right)} \label{eq:approx_softmax_1}  \\
        &= \frac{\boldsymbol{\phi}\left(\bq_t\right)^\top\sum_i \boldsymbol{\phi}\left(\bk_i\right)\otimes\bv_i}
        {\boldsymbol{\phi}\left(\bq_t\right) ^\top \sum_j \boldsymbol{\phi}\left(\bk_j\right)},\label{eq:approx_softmax_2}
\end{align}
\end{subequations}
where $\otimes$ refers to the outer product.
Recent works have shown that kernel-based linear attentions perform favorably against the original softmax attention 
on machine translation~\cite{peng2020random} and protein sequence modeling~\cite{choromanski2021rethinking}. However, although the complexity is reduced into linear, the intensive floating-point multiplications in Eq.~(\ref{eq:approx_softmax_2}) still consume a large amount of energy, which can quickly drain the batteries on mobile/edge platforms.

\subsection{Binary Quantization}
\label{sec:binary_quantization}
Following~\cite{rastegari2016xnor}, binary quantization typically estimates the full-precision $\bu \in \mathbb{R}^n$ using a binary $\hat{\bu} \in \{+1, -1\}^n$ and a scaling factor $\alpha \in \mathbb{R}^+$ such that $\bu \approx \alpha \hat{\bu}$ holds. To find an accurate estimation, existing methods \revise{\cite{rastegari2016xnor, martinez2020training, lin2017towards, bai2021binarybert, qin2022bibert}} minimize the quantization error as
\begin{equation}
    \label{eq:binary_estimation}
    \alpha^{*},\hat{\bu}^{*} = \arg\min \| \bu - \alpha \hat{\bu} \|.
\end{equation}
By solving Problem~(\ref{eq:binary_estimation}), we have $\hat{\bu} = \mathrm{sign}(\bu)$ and $\alpha = \frac{1}{n} \| \bu \|_{\ell 1}$,
where $\mathrm{sign}(u)$ returns 1 if $u \ge 0$ and -1 if $u < 0$. Since the sign function is non-differentiable, the straight-through estimator (STE)~\cite{bengio2013estimating} is applied to approximate the gradient such as using the gradient of hard $\mathrm{tanh}$~\cite{hubara2017quantized} or piecewise polynomial function~\cite{Liu_2018_ECCV}.

\vspace{-2mm}
\section{Proposed Method}
To reduce the energy consumption of self-attention, one may perform binary quantization~\cite{qin2022bibert,bai2021binarybert} on the queries $\{\bq_t\}^N_{t=1}$ and keys $\{\bk_t\}^N_{t=1}$. In this case, we can replace most of the energy-expensive multiplications with the energy-efficient bit-wise operations.
However, existing binary quantization methods only focus on minimizing the quantization error between the original full-precision values and the binary ones as in Eq.~(\ref{eq:binary_estimation}), which fails to preserve the pairwise semantic similarity between different tokens in attention, leading to performance degradation.

Note that the attention can be seen as applying kernel smoother over pairwise tokens where the kernel scores denote the similarity of the token pairs, as mentioned in Section~\ref{sec:linear_attention}. Motivated by this, we propose a new binarization method that applies kernelized hashing with Gaussian RBF to map the original high-dimensional queries/keys to low-dimensional similarity-preserving binary codes in Hamming space. 
The proposed framework, which we dub \textit{EcoFormer}, is depicted in Figure~\ref{fig:framework} (c).
To maintain the semantic similarity in attention, we learn the hash functions in a self-supervised manner. By exploiting the associative property of the linear dot-product between binary codes and the equivalence between the code inner products (\ie, Hamming affinity) and the Hamming distances, we are able to approximate the self-attention in linear time with low energy cost. In the following, we first introduce the kernelized hashing attention in Section~\ref{sec:kernelized_hashing_attention} and then show how to learn the hash functions in a self-supervised way in Section~\ref{sec:self_supervised_hashing}.

\subsection{Kernelized Hashing Attention}
\label{sec:kernelized_hashing_attention}
Before applying hash functions, we let the queries $\{\bq_t\}^N_{t=1}$ and keys $\{\bk_t\}^N_{t=1}$ to be identical following~\cite{reformer,lu2021soft}. In this way, 
we can then apply kernelized hash functions $H : \mathbb{R}^{D_p} \mapsto \{1, -1\}^{b} $ without explicitly applying transformation $\boldsymbol{\phi}(\cdot)$ mentioned in Section~\ref{sec:linear_attention} to map ${\bq}_i$ and ${\bk}_j$ into $b$-bit binary codes $H({\bq}_i)$ and $H({\bk}_j)$, respectively (see Section~\ref{sec:self_supervised_hashing}).
In this case, the Hamming distance between them can be defined as 
\begin{equation}
{\begin{array}{ll}
    \label{eq:hamming_distance}
     \mD\left(H({\bq}_i), H({\bk}_j)\right)= \sum_{r=1}^b \mathbbm{1} \{ H_r({\bq}_i) \ne H_r({\bk}_j) \},
\end{array}}
\end{equation}
where $H_r(\cdot)$ is the $r$-th bit of the binary codes; $\mathbbm{1}\{A\}$ is an indicator function that returns 1 if $A$ is satisfied and otherwise returns 0. With $\mD\left(H({\bq}_i), H({\bk}_j)\right)$, the codes inner product between $H({\bq}_i)$ and $H({\bk}_j)$ can be formulated as
\begin{equation}
    \begin{aligned}
        \label{eq:relation_inner_product_hamming_distance}
         H({\bq}_i)^{\top} H({\bk}_j) &= \sum_{r=1}^b \mathbbm{1} \{ H_r({\bq}_i) = H_r({\bk}_j) \} - \sum_{r=1}^b \mathbbm{1} \{ H_r({\bq}_i) \ne H_r({\bk}_j) \} \\
        &= b - 2 \sum_{r=1}^b \mathbbm{1} \{ H_r({\bq}_i) \ne H_r({\bk}_j) \}
        = b - 2 \mD\left(H({\bq}_i), H({\bk}_j)\right).
    \end{aligned}
\end{equation}
Importantly, Eq.~(\ref{eq:relation_inner_product_hamming_distance}) shows the equivalence between the Hamming distance and the codes inner product since there is a one-to-one correspondence. 
By substituting with the hashed queries and keys in Eq.~(\ref{eq:approx_softmax_1}), we can approximate the self-attention as
\begin{equation}
\begin{aligned}
    \label{eq:hashing_attention_vanilla}
    \mathrm{Attention}(\bq_t, \{\bk_i\}, \{\bv_i\}) 
    &\approx     \sum_i\frac{ H( {\bq}_t )^{\top} H({\bk}_i)  \bv_i}
    {\sum_j H( {\bq}_t ) ^\top H({\bk}_j) }.
\end{aligned}
\end{equation}
Note that $H( {\bq}_t )^\top H({\bk}_j) \in [-b, b]$. To avoid zero in denominator, we introduce a bias term $2^c$ to each inner product so that $H( {\bq}_t )^\top H({\bk}_j) + 2^c > 0$, having no effect on the similarity measure. Here, we can simply set $c$ to $\lceil \log_2 (b+1) \rceil$ where $\lceil u \rceil$ returns the least integer greater than or equal to $u$.
Using the associative property of matrix multiplication, we approximate the self-attention as
\begin{equation}
    \begin{aligned}
        \label{eq:hashing_attention}
        \mathrm{Attention}(\bq_t, \{\bk_i\}, \{\bv_i\}) 
        &\approx     \sum_i\frac{ \left( H( {\bq}_t )^{\top} H({\bk}_i) + 2^c\right)  \bv_i}
        {\sum_j \left( H( {\bq}_t ) ^\top H({\bk}_j) + 2^c \right) } \\
        &= \frac{ H( {\bq}_t )^{\top} \sum_i H({\bk}_i) \otimes \bv_i + \sum_i 2^c\bv_i ^\top }{H( {\bq}_t ) ^\top \sum_j H({\bk}_j) + {2^c}N}.
    \end{aligned}
\end{equation}
In practice, the multiplications between the binary codes and the full-precision values in Eq.~(\ref{eq:hashing_attention}) can be replaced by simple additions and subtractions, which greatly reduce the computational overhead in terms of on-chip energy footprint. Moreover, the multiplications with a powers-of-two $2^c$ can also be implemented by efficient \textit{bit-shift} operations. As a result, the only multiplications come from the element-wise divisions between the numerator and denominator.

\subsection{Self-supervised Hash Function Learning}
\label{sec:self_supervised_hashing}
Given queries $\mQ=\{ \bq_1, \dots, \bq_N \} \subset \mathbb{R}^{D_p}$,
we seek to learn a group of hash functions $h : \mathbb{R}^{D_p} \mapsto \{1,-1\}$. Instead of explicitly applying the transformation function $\boldsymbol{\phi}(\cdot)$ mentioned in Section~\ref{sec:linear_attention}, we compute the hash functions with a kernel function $\kappa(\bq_i, \bq_j) : \mathbb{R}^{D_p} \times \mathbb{R}^{D_p} \mapsto \mathbb{R}$. Given $\bQ = \left[ \bq_1, \cdots, \bq_N \right]^{\top} \in \mathbb{R}^{N \times D_p}$,
we randomly sample $m$ queries ${\bq}_{(1)}, \dots, {\bq}_{(m)}$ from $\mQ$ as support samples following the kernel-based supervised hashing (KSH)~\cite{liu2012supervised} and define a hash function $h$ as
\begin{equation}
    \label{eq:prediction_f}
    {\begin{array}{ll}
    h({\bQ}) = \mathrm{sign} \left( \sum_{j=1}^m \left( \kappa\left({\bq}_{(j)}, {\bQ}\right) - \mu_j \right) a_j \right)  = \mathrm{sign} \left( \bg({\bQ}) \ba \right),
    \end{array}}
\end{equation}
where $\ba = [a_1, \cdots, a_m]^{\top}$ is the weight of $h$, $\mu_j = \frac{1}{n} \sum_{i=1}^N \kappa\left({\bq}_{(j)}, {\bq}_i \right)$ is to normalize the kernel function to zero-mean, and $\bg: \mathbb{R}^{D_p} \mapsto \mathbb{R}^m$ is a mapping defined by {$\bg({\bQ}) = \left[ \kappa\left({\bq}_{(1)}, {\bQ}\right) - \mu_1, \dots, \kappa({\bq}_{(m)}, {\bQ}) - \mu_m \right] \in \mathbb{R}^{N \times m}$.}
Then, we define the kernelized hash function $H(\cdot)$ as
\begin{equation}
H({\bQ})= \left[ h_1(\bQ), \cdots, h_b(\bQ) \right] = \left[ \begin{array}{c}
     h_1(\bq_1), \cdots, h_b(\bq_1) \\
     \cdots \cdots \\
     h_1(\bq_N), \cdots, h_b(\bq_N)
\end{array}
\right] =
\mathrm{sign}\left(\bg(\bQ) \bA\right),
\end{equation}
where 
$\bA=\left[\ba_1,\cdots,\ba_b\right] \in \mathbb{R}^{m \times b}$, {and $h_r({\bQ})=\mathrm{sign}\left( \bg(\bQ) \ba_r \right)$ is the hash function for the $r$-th bit.}

To guide the learning of the binary codes, we hope that similar token pairs will have the minimal Hamming distance while dissimilar token pairs will have the maximal distance. 
Nevertheless, directly optimizing the Hamming distance is difficult due to the non-convex and non-smooth formulation in Eq.~(\ref{eq:hamming_distance}). Utilizing the equivalence between the code inner products and the Hamming distances in Eq.~(\ref{eq:relation_inner_product_hamming_distance}), we instead conduct optimization based on the Hamming affinity to minimize the reconstruction error as
\begin{equation}
    \label{eq:hashing_objective}
    {\begin{array}{ll}
    \min_{\bA} \left\| H({\bQ}) H({\bQ})^{\top} - b\bY \right\|_F^2 = \min_{\bA} \left\| \sum_{r=1}^b h_r({\bQ}) h_r({\bQ})^{\top} - b\bY \right\|_F^2,
    \end{array}}
\end{equation}
where $\left\| \cdot \right\|_F$ is the Frobenius norm and $\bY \in \mathbb{R}^{N \times N}$ is the target Hamming affinity matrix. 
To preserve the similarity relations between queries and keys, we use the attention scores as the \textit{self-supervised} information to construct $\bY$.
Let $\mS$ and $\mU$ be the similar and dissimilar pairs of tokens. We obtain $\mS$ and $\mU$ by selecting the token pairs with the Top-$l$ largest and smallest attention scores. We then construct pairwise labels $\bY$ as
\begin{equation}
    \label{eq:pairwise_label}
    \begin{array}{ll}
    \bY_{i j}= 
    \begin{cases}1, & \left(\bq_{i}, \bq_{j}\right) \in \mS \\ -1, & \left(\bq_{i}, \bq_{j}\right) \in \mU \\ 0, & \text{otherwise}.
    \end{cases}
    \end{array}
\end{equation}
However, Problem~(\ref{eq:hashing_objective}) is NP-hard. To solve it efficiently, we adapt discrete cyclic coordinate descent to learn binary codes sequentially. 
Specifically, we only solve $\ba_r$ once the previous $\ba_1, \cdots, \ba_{r-1}$ have been optimized. {Let $\hat{\bY}_{r-1} = b\bY - \sum_{t=1}^{r-1} h_t({\bQ}) h_t({\bQ})^{\top}$ be the residual matrix}, where $\hat{\bY}_0 = b\bY$. Then, we can minimize the following objective to obtain $\ba_r$
\begin{equation}
    {\begin{aligned}
        \label{eq:final_objective}
        \begin{array}{ll}
        \min_{\ba_r}&\left\| h_r({\bQ}) h_r({\bQ})^{\top} - \hat{\bY}_{r-1} \right\|_F^2 
        =\min_{\ba_r}- 2h_r({\bQ})^{\top} \hat{\bY}_{r-1} h_r({\bQ}) + C,
        \end{array}
    \end{aligned}}
\end{equation}
{where $C=\left(h_r({\bQ})^{\top} h_r({\bQ})\right)^2 + \mathrm{tr}\left(\hat{\bY}_{r-1}\right)$ is a constant.}
Note that $\hat{\bY}_{r-1}$ is a symmetric matrix. Therefore, Problem~(\ref{eq:final_objective}) is a standard binary quadratic programming problem, which can be efficiently solved by many existing methods, such as the LBFGS-B solver \cite{zhu1997algorithm} and block graph cuts~\cite{lin2014fast}. To learn $\ba_r$ in conjunction with network parameters,
we propose to solve Problem~(\ref{eq:final_objective}) using the gradient-based methods. 
For the non-differentiable sign function, we use STE~\cite{bengio2013estimating} to approximate the gradient using hard $\mathrm{tanh}$ as mentioned in Section~\ref{sec:binary_quantization}. Note that learning the hash functions for each epoch is computationally expensive yet unnecessary. We only learn the hash functions per $\tau$ epoch.

\vspace{-2mm}
\section{Experiments}
\label{sec:experiments}
\begin{table}[htp]
\centering
\renewcommand\arraystretch{1.2}
\caption{Main results on ImageNet-1K. The number of multiplications, additions, as well as \revise{on-chip} energy consumption are calculated based on an image of resolution $224\times224$. The throughput is measured with a mini-batch size of 32 on a single NVIDIA RTX 3090 GPU.}
\label{tab:main_results}
\scalebox{0.80}{
\begin{tabular}{l|c|ccccc}
Model                         & Method & \#Mul. (B) & \#Add. (B) & Energy (B pJ) & Throughput (images/s)  & Top-1 Acc. (\%) \\ \shline
\multirow{2}{*}{PVTv2-B0~\cite{pvtv2}}    & MSA    & 2.02     & 1.99     & 9.25  & 850        & 70.77    \\
 & \textbf{Ours} & \textbf{0.54} & \textbf{0.56} & \textbf{2.49}  & \textbf{1379} & \textbf{70.44}   \\ \hline
\multirow{2}{*}{PVTv2-B1}    & MSA    & 5.02     & 5.00     & 23.07 & 621    & 78.83    \\
 & \textbf{Ours} & \textbf{2.03} & \textbf{2.09} & \textbf{9.39} & \textbf{874} & \textbf{78.38}   \\ \hline
\multirow{2}{*}{PVTv2-B2}    & MSA    & 8.64     & 8.60     & 39.71 & 404  & 81.82    \\
 & \textbf{Ours} & \textbf{3.85} & \textbf{3.97} & \textbf{17.82} & \textbf{483} & \textbf{81.28}   \\ \hline
\multirow{2}{*}{PVTv2-B3}    & MSA    & 11.86     & 11.82     & 54.56  & 310              & 82.26    \\
 & \textbf{Ours} & \textbf{6.54}     & \textbf{6.72}     & \textbf{30.25} & \textbf{325}       & \textbf{81.96}    \\   \hline
\multirow{2}{*}{PVTv2-B4}    & MSA    & 15.97     & 15.93     & 73.43     & 247           & 82.42    \\ 
 & \textbf{Ours} & \textbf{9.57}     & \textbf{9.82}     & \textbf{44.25}  & \textbf{249}          & \textbf{81.90}     \\ \hline
\multirow{2}{*}{Twins-SVT-S~\cite{twins}} & MSA    & 5.96     & 5.91     & 27.36 & 426 & 81.66    \\
 & \textbf{Ours} & \textbf{2.72} & \textbf{2.81} & \textbf{12.59} & \textbf{576} & \textbf{80.22} \\
\end{tabular}
}
\vspace{-3mm}
\end{table} 

\subsection{Comparisons on ImageNet-1K}
%\vspace{-3mm}
\label{sec:results_imagenet}
To investigate the effectiveness of the proposed method, we conduct experiments on ImageNet-1K~\cite{krizhevsky2012imagenet}, a large-scale image classification dataset that contains $\sim$1.2M training images from 1K categories and 50K validation images. 
We compare our kernelized hashing attention with standard MSA by adapting the two attention approaches into two popular vision Transformer frameworks PVTv2~\cite{pvtv2} and Twins~\cite{twins}. We measure model performance by the Top-1 accuracy. Furthermore, as FLOPs cannot accurately reflect the computational cost in our proposed method, we measure the model complexity by the number of multiplications and additions, separately, as done in \cite{shu2021adder}. \revise{Specifically, we calculate FLOPs following~\cite{wang2021pyramid}, where we count the multiply-accumulate operations for all layers. In this case, each multiply-accumulate operation consists of an addition and a multiplication. We also count the multiplications in the scaling operations.} 
Moreover, we report the \revise{on-chip} energy consumption according to Table~\ref{table:energy_area_cost} and the throughput with a mini-batch size of 32 on a single NVIDIA RTX 3090 GPU.

\noindent\textbf{Implementation details.}
All training images are resized to $256 \times 256$, and $224 \times 224$ patches are randomly cropped from an image or its horizontal flip, with the per-pixel mean subtracted. To obtain the MSA baselines, we first replace the original attention layers in PVTv2~\cite{pvtv2} and Twins~\cite{twins} with standard MSAs and initialize the models with the pretrained weights on ImageNet-1K. Next, we finetune each model on ImageNet-1K with 100 epochs. 
Based the pretrained MSA weights, we then apply our approach to each model and finetune on ImageNet-1K with 30 epochs. All models in this experiment are trained on 8 V100 GPUs with a total batch size of 256. We set the initial learning rate to $2.5\times10^{-5}$ for PVTv2 and $1.25\times10^{-4}$ for Twins. We use AdamW optimizer~\cite{adamw} with a cosine decay learning rate scheduler. All other hyperparameters are the same as in PVTv2. Also note that recent hierarchical ViTs~\cite{pvtv2,liu2021swin,twins} have multiple stages to incorporate pyramid feature maps. At the last stage, they usually apply standard MSAs due to the comparably low-resolution feature maps. This design is also adopted in PVTv2 and Twins. Therefore, we follow the common practice and do not modify the attention layers at the last stage. For the hash functions learning, we set the number of support samples $m$ and update interval $\tau$ to 25 and 30, respectively. The hyper-parameter $l$ in constructing pairwise labels $\bY$ is set to 10. We set the hash bit $b$ to 16.

\begin{figure}[t!]
	\centering
	\includegraphics[width=0.8\linewidth]{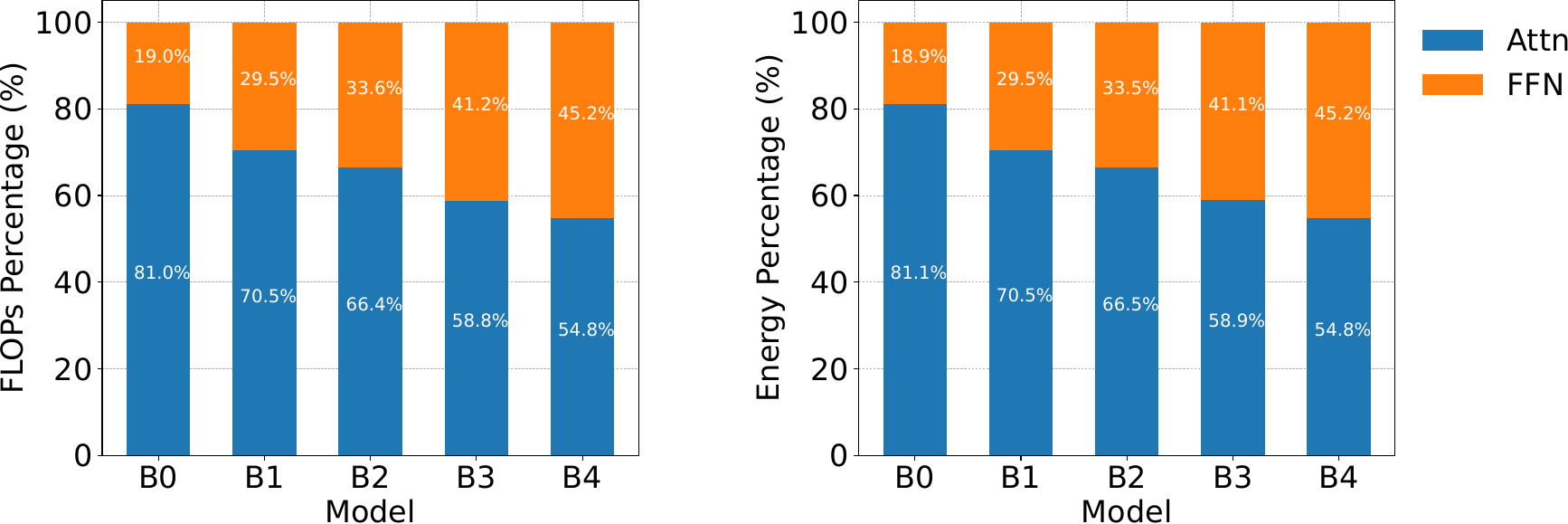}
	\caption{\revise{FLOPs and on-chip energy footprint percentage of attention layers (Attn) and feed-forward layers (FFN) in different variants of PVTv2 with standard MSAs. 
 % ``Attn'' and ``FFN'' denote the attention layers and the feed-forward layers, respectively. 
 For a bigger model, FFN takes a larger proportion of the computational cost and \revise{on-chip} energy footprint.}}
        \vspace{-3mm}
	\label{fig:combine_ops_energy}
\end{figure}

\noindent\textbf{Results analysis.} 
We report the results in Table~\ref{tab:main_results}.
\revise{In general, our baseline MSA has more multiplications than additions. In contrast, our EcoFormer replaces most of the floating-point multiplications in attention with simple additions. Therefore, there are more additions than multiplications in our EcoFormer.}
Compared to MSA, our method achieves lower computational complexity, less energy consumption and higher throughput with comparable performance. For example, based on PVTv2-B0, our method saves around 73\% multiplications and 72\% additions, as well as reducing 73\% on-chip energy consumption, which demonstrates the energy-efficiency of our approach. With more efficient accumulation implementation, the throughput of our EcoFormer can be further improved. Besides, \revise{a larger model comes with a larger proportion of computational and \revise{on-chip} energy cost dominated by FFNs, as shown in Figure~\ref{fig:combine_ops_energy}.} 
In this case, as our approach focuses on the attention layers, the energy-saving from larger models is comparably less than smaller models (\eg, PVTv2-B0 vs. \revise{PVTv2-B4}). Nonetheless, we still reduce significant computational cost and \revise{on-chip} energy consumption. For example, on \revise{PVTv2-B4}, we save around \revise{40\% on-chip} energy consumption. Compared with PVTv2, the performance drop of our method is slightly larger on Twins. One possible reason is that our method may be sensitive to the conditional positional encodings in Twins.

\vspace{-1mm}
\subsection{Comparisons on Long Range Arena}
To evaluate the performance of different efficient attentions under long-context scenarios, we train our EcoFormer on two tasks, \textbf{Text} and \textbf{Retrieval} from the Long Range Arena (LRA) benchmark~\cite{tay2020long} following the settings of~\cite{zhu2021long}. Our implementations are based on the released code of~\cite{xiong2021nystromformer}. We use the same hyper-parameters $m$, $\tau$ and $b$ as in ImageNet-1K experiments. We show the results in Table~\ref{tab:long_range_arena}. From the table,  our EcoFormer achieves comparable performance with much lower \revise{on-chip} energy consumption. For example, on \textbf{Text}, compared with MSA, our method saves around 94.6\% multiplications and 93.7\% additions as well as 94.5\% \revise{on-chip} energy consumption, which is more efficient than existing attention mechanisms. 

\vspace{-5mm}
\begin{table}[htp]
\centering
\caption{Comparisons of different methods on Long Range Arena (LRA). We report the classification accuracy (\%) for \textbf{Text} as well as \textbf{Retrieval} and average accuracy across two tasks. $^{*}$ denotes that we obtain the results from the original paper.}
\renewcommand\arraystretch{1.2}
\label{tab:long_range_arena}
\scalebox{0.83}{
\begin{tabular}{l|ccccccc}
Method      & \#Mul. (B) & \#Add. (B) & Energy (B pJ) & Text (4K) & Retrieval (4K) & Average \\ \shline
Transformer & 4.63 & 4.57 & 21.25 & 64.87 & 79.62 & 72.25 \\
Performer~\cite{choromanski2021rethinking} & 0.83 & 0.84 & 3.83 & 64.82 & 79.08 & 71.95 \\
Linformer~\cite{linformer} & 0.81 & 0.81 & 3.74 & 57.03 & 78.11 & 67.57 \\
Reformer~\cite{reformer} & 0.54 & 0.54 & 2.49 & 65.19 & 79.46 & 72.33 \\
Combiner$^{*}$~\cite{ren2021combiner} & 0.51 & 0.51 & 2.34 & 64.36 & 56.10 & 60.23 \\
\textbf{EcoFormer} & \textbf{0.25} &  \textbf{0.29} & \textbf{1.17} & 64.79 & 78.67 & 71.73 \\
\end{tabular}
}
\vspace{-5mm}
\end{table}

\subsection{Ablation Study}
In this section, we evaluate the effectiveness of our EcoFormer by comparing it with different binarization approaches and efficient attention mechanisms. 
By default, we train each model from scratch on CIFAR-100 with 2 GPUs for 300 epochs. The total batch size is 64. The initial learning rate is $6.25 \times 10^{-6}$. For the hash functions learning, we set update interval $\tau$ to 300. All the other hyperparameters are the same as in ImageNet-1K experiments.

\begin{table}[htp]
\centering
\vspace{-4mm}
\renewcommand\arraystretch{1.2}
\caption{Performance comparisons with different binarization methods on CIFAR-100.}
\label{tab:quant_vs_hashing}
\scalebox{0.83}{
\begin{tabular}{l|l|ccccc}
Model & Method             & \#Mul. (B)     & \#Add. (B)     & Energy (B pJ)  & Top-1 Acc. (\%)          \\ \shline
\multirow{3}{*}{PVTv2-B0}    & FP-EcoFormer & 0.94 & 0.94 & 4.33  & 70.78 \\
                             & Bi-EcoFormer      & 0.63 & 0.83 & 3.09  & 70.06 \\ 
          & \textbf{EcoFormer} & \textbf{0.54} & \textbf{0.56} & \textbf{2.49}  & \textbf{71.23} \\ \hline
\multirow{3}{*}{Twins-SVT-S} & FP-EcoFormer & 5.96 & 5.91 & 27.36 & 80.04 \\
                             & Bi-EcoFormer      & 3.01 & 3.59 & 14.38 & 80.04 \\
          & \textbf{EcoFormer} & \textbf{2.72} & \textbf{2.81} & \textbf{12.58} &  \textbf{80.31}
\end{tabular}
}
% \vspace{-2mm}
\end{table}

\noindent\textbf{Quantization vs. hashing.}
To investigate the effect of different binarization methods, we compare our \methodshortname with the following methods: \textbf{FP-\methodshortname}: Based on \methodshortname, we do not binarize queries and keys in attentions.
\textbf{Bi-\methodshortname}: Relying on \methodshortname, we use the same binary quantization ~\cite{hubara2017quantized} as BinaryBERT~\cite{bai2021binarybert} and BiBERT~\cite{qin2022bibert} to obtain binarized queries and keys instead of our proposed hash functions. For fair comparisons, the attention operations in the compared method are in linear complexity.
We apply different methods to PVTv2-B0 and Twins-SVT-S and report the results in Table~\ref{tab:quant_vs_hashing}. 
We observe that our \methodshortname consistently outperforms Bi-\methodshortname on different frameworks. For example, based on PVTv2-B0, our \methodshortname surpasses Bi-\methodshortname by 1.17\% in terms of the Top-1 accuracy. Compared with binary quantization, our proposed self-supervised hash functions preserve the pairwise similarity of attention, leading to better performance. Moreover, our \methodshortname does not need to explicitly compute transformation $\boldsymbol{\phi}(\cdot)$ as in Eq.~(\ref{eq:approx_softmax_2}).
Therefore, the energy cost of our \methodshortname is lower than Bi-\methodshortname.

\begin{table}[htp]
\centering
\vspace{-4mm}
\renewcommand\arraystretch{1.2}
\caption{Performance comparisons with different hash functions regarding PVTv2-B0 on CIFAR-100.}
\label{tab:diff_hashing}
\scalebox{0.83}{
\begin{tabular}{l|lccccc}
Method & \#Mul. (B)     & \#Add. (B)     & Energy (B pJ)  & Top-1 Acc. (\%)          \\ \shline
LSH-EcoFormer & 0.68 & 0.69 & 3.12 & 70.18 \\
KLSH-EcoFormer & 0.54 & 0.56 & 2.49 & 70.66 \\
\textbf{EcoFormer} & \textbf{0.54} & \textbf{0.56} & \textbf{2.49} & \textbf{71.23}
\end{tabular}
}
\vspace{-2mm}
\end{table}

\noindent\textbf{Effect of different hash functions.}
To investigate the effect of different hash functions, we include the following methods for comparisons: \textbf{LSH-\methodshortname}: Relying on \methodshortname, we use Locality-Sensitive Hashing (LSH)~\cite{datar2004locality} rather than our proposed kernelized hash function. \textbf{KLSH-\methodshortname}: Based on \methodshortname, we replace the proposed hash function with Kernelized Locality-Sensitive Hashing (KLSH)~\cite{kulis2011kernelized}. We report the results in Table~\ref{tab:diff_hashing}. 
We can observe that KLSH-\methodshortname outperforms LSH-\methodshortname by 0.48\% in terms of the Top-1 accuracy with less \revise{on-chip} energy consumption.
The reason can be attributed to that LSH is based on random linear projection, which can not deal with the non-linear softmax attention well.
Critically, our \methodshortname further improves the performance by 0.57\% on the Top-1 accuracy. Compared with random hashing in KLSH, our \methodshortname learns the hash functions with additional self-supervised information. Therefore, our learned binary codes are better at preserving the token similarity.

\vspace{-5mm}
\begin{table}[htp]
\centering
\caption{Comparison with other efficient attention methods regarding PVTv2-B0~\cite{pvtv2} on CIFAR-100. }
\renewcommand\arraystretch{1.2}
\label{tab:compared_diff_attn}
\scalebox{0.82}{
\begin{tabular}{l|ccccc}
Method      & \#Mul. (B) & \#Add. (B) & Energy (B pJ) & Top-1 Acc. (\%)         \\ \shline
Transformer   & 2.02          & 1.99          & 9.25          & 71.44 \\
Performer~\cite{choromanski2021rethinking}     & 0.94          & 0.94          & 4.33          & 70.78 \\
Linformer~\cite{linformer}     & 0.69          & 0.69          & 3.18            & 71.17 \\
Reformer~\cite{reformer}      & 1.62          & 1.63          & 7.44           & 70.56 \\
\textbf{EcoFormer} & \textbf{0.54} & \textbf{0.56} & \textbf{2.49} & \textbf{71.23}
\end{tabular}
}
\vspace{-5mm}
\end{table}

\begin{wraptable}{R}{0.5\textwidth}
\vspace{-5mm}
\begin{minipage}{0.5\textwidth}
\centering
\caption{Latency and energy comparisons with different attention methods. We measure the latency and energy of an attention layer with a batch size of 16, a sequence length of 3,136 and an embedding dimension of 32 on a BitFusion~\cite{sharma2018bit} simulator.
}
% \vspace{-1mm}
\label{tab:bitfusion}
\renewcommand\arraystretch{1.1}
\scalebox{0.8}{
\begin{tabular}{l|ccccc}
Method      & Latency (ms) & Energy (pJ) \\ \shline
Transformer   & 0.0036 & 85,692.18 \\
Performer~\cite{choromanski2021rethinking} & 0.0019 & 41,113.64 \\
Linformer~\cite{linformer} & 0.0018 & 45,770.61 \\
Reformer~\cite{reformer} & 0.0024 & 57,305.47 \\
\textbf{EcoFormer} & \bf{0.0010} & \bf{24,990.75} \\
\end{tabular}
}
\end{minipage}
% \vspace{-2mm}
\end{wraptable}

\paragraph{Comparing with other efficient attention mechanisms.}
To compare EcoFormer with different attention mechanisms, we conduct experiments on CIFAR-100 based on PVTv2-B0 in Table~\ref{tab:compared_diff_attn}. In our experiments, we directly replace the attention layers with each compared method in PVTv2-B0~\cite{pvtv2}. In general, compared to other efficient attention mechanisms, EcoFormer saves more computations and reduces more \revise{on-chip} energy consumption while achieving better performance. Particularly, benefiting from the multiplication-saving operations and low-dimensional binary queries and keys, EcoFormer saves more \revise{on-chip} energy than Performer. Also note that since the proposed kernelized hash function $H(\cdot)$ does not need to explicitly apply transformation $\boldsymbol{\phi}(\cdot)$ to the queries and keys as in Eq.~(\ref{eq:approx_softmax_2}),
EcoFormer simultaneously reduces more multiplications and additions than Performer.
Besides, Linformer achieves competitive results. However, as the size of the learnable low-rank projection parameters depends on the length of the input sequence, Linformer is not scalable to different image resolutions, whereas EcoFormer with sufficient bits is agnostic to the sequence length.

\paragraph{Latency and energy on BitFusion~\cite{sharma2018bit}.} 
To show the actual energy consumption and latency, we test different methods on a simulator of BitFusion, a bit-flexible microarchitecture synthesized in 45 nm technology. From Table~\ref{tab:bitfusion}, EcoFormer shows much lower latency and \revise{on-chip} energy than the other efficient attention methods, which further verifies the advantage of EcoFormer.

\paragraph{\revise{Effect of training from scratch on ImageNet-1K.}} 
\revise{To explore the effect of training from scratch, we apply EcoFormer to PVTv2-B0 and PVTv2-B1. We follow the experimental settings mentioned in Section~\ref{sec:results_imagenet} except that we train the model from scratch with 300 epochs. The initial learning rate is set to $2.5 \times 10^{-4}$. From Table~\ref{tab:effect_training_from_scrach}, EcoFormer achieves comparable performance while significantly reducing the computational complexity and on-chip energy consumption. The accuracy drop from discretization comes from the gradient approximation for the non-differentiable $\rm{sign}$ function, which can be mitigated by more advanced optimization methods, such as regularization \cite{ding2019regularizing}, knowledge distillation \cite{mishra2017apprentice, qin2022bibert}, relaxed optimization \cite{hou2018loss, bai2019proxquant}, appending full-precision branches \cite{Liu_2018_ECCV, martinez2020training}, \etc}

\vspace{-4mm}
\begin{table}[htp]
\centering
\renewcommand\arraystretch{1.2}
\caption{\revise{Performance comparisons of different methods on ImageNet-1K. All the models are trained from scratch. The number of multiplications, additions, and on-chip energy consumption are calculated based on an image of resolution $224\times224$.}}
\label{tab:effect_training_from_scrach}
\scalebox{0.85}{
\revise{\begin{tabular}{l|c|ccccc}
Model                         & Method & \#Mul. (B) & \#Add. (B) & Energy (B pJ) & Top-1 Acc. (\%) \\ \shline
\multirow{2}{*}{PVTv2-B0}    & MSA    & 2.02     & 1.99     & 9.25                & 69.72    \\ 
& \textbf{Ours} & \textbf{0.54} & \textbf{0.56} & \textbf{2.49} & \textbf{68.70} \\
\hline
\multirow{2}{*}{PVTv2-B1}    & MSA    & 5.02     & 5.00     & 23.07                & 78.34    \\ 
& \textbf{Ours} & \textbf{2.03} & \textbf{2.09} & \textbf{9.39} & \textbf{77.49} \\
\end{tabular}}
}
\vspace{-4mm}
\end{table} 

\paragraph{\revise{Effect of different $m$.}} 
\revise{To investigate the effect of different numbers of support samples $m$, we train \methodshortname with different $m$ based on PVTv2-B0. We report the results on CIFAR-100 in Table~\ref{tab:effect_m}. As we increase $m$, the performance becomes better along with the increase in on-chip energy consumption. For example, the model obtained with $m=15$ outperforms that of $m=10$ by 0.19\% on the Top-1 accuracy with little additional energy cost. We speculate that, with more support samples, we can capture more accurate statistics in Eq.~(\ref{eq:prediction_f}) and hence lead to better performance. Since our EcoFormer achieves the best performance with $m=25$, we use it by default in our experiments.}

\vspace{-5mm}
\begin{table}[h]
\centering
\renewcommand\arraystretch{1.2}
\caption{\revise{Performance comparisons with different \#support samples $m$. We report the results of PVTv2-B0 on CIFAR-100.}}
\label{tab:effect_m}
\scalebox{0.85}{
\revise{\begin{tabular}{c|ccccc}
$m$ &  \#Mul. (B) & \#Add. (B) & Energy (B pJ) & Top-1 Acc. (\%) \\
\shline
10 & 0.53 & 0.55 & 2.46 & 70.73\\
15 & 0.53 & 0.56 & 2.47 & 70.92\\
20 & 0.53 & 0.56 & 2.48 & 70.81\\
25 & 0.54 & 0.56 & 2.49 & \textbf{71.23}\\
\end{tabular}}}
\vspace{-4mm}
\end{table}

\vspace{-2mm}
\section{Conclusion and Future Work}
\vspace{-2mm}
\label{sec:conclusion}
In this paper, we have presented a novel energy-saving attention mechanism with linear complexity to save the vast majority of multiplications from a new binarization perspective, making the deployment of Transformer models at scale feasible on edge devices.
We are inspired by the fact that conventional binarization methods are built upon statistical quantization error minimization without considering to preserve the pairwise similarity relations between tokens. To this end, we customize binarization to softmax attention by mapping the original token features into compact binary codes in Hamming space using a set of kernel-based hash functions, where the similarity can be measured by codes dot product. 
The hash functions for queries/keys are learned to encourage the Hamming affinity of a token pair to be close to the target obtained from the attention scores, in a self-supervised way. 
Extensive experiments have demonstrated that EcoFormer saves significant \revise{on-chip} energy footprint while achieving comparable performance with standard attentions on ImageNet-1K, Long Range Arena and CIFAR-100. 
In terms of the future work, we can further binarize the value vectors in attention, multi-layer perceptrons and non-linearities in Transformer to make it fully binarized for more significant \revise{on-chip} energy-saving. We may also extend EcoFormer to other NLP tasks such as machine translation and 
speech analysis tasks to make it more impactful to wider communities. 

\noindent\textbf{Limitations and societal impact.}
We have shown that EcoFormer is more energy-efficient compared to the standard attention. However, in practice, the addition operations between binary codes and floating-point numbers will require specialized GPU kernels (\eg, customized CUDA operators) for further acceleration. \revise{Moreover, for the short sequence scenario, our EcoFormer suffers from severe performance drop due to the limited representational capability.}
Our work potentially brings some negative societal impacts that training large Transformer models requires extensive computations, resulting in financial and environmental costs. A promising solution is to jointly optimize training and inference efficiency.

\bibliographystyle{abbrv}
{
    \small
	\bibliography{reference}

\begin{thebibliography}{10}

\bibitem{andoni2015practical}
A.~Andoni, P.~Indyk, T.~Laarhoven, I.~Razenshteyn, and L.~Schmidt.
\newblock Practical and optimal lsh for angular distance.
\newblock In {\em NeurIPS}, volume~28, pages 1225--1233, 2015.

\bibitem{bai2021binarybert}
H.~Bai, W.~Zhang, L.~Hou, L.~Shang, J.~Jin, X.~Jiang, Q.~Liu, M.~Lyu, and
  I.~King.
\newblock Binarybert: Pushing the limit of bert quantization.
\newblock In {\em ACL}, pages 4334--4348, 2021.

\bibitem{bai2019proxquant}
Y.~Bai, Y.-X. Wang, and E.~Liberty.
\newblock Proxquant: Quantized neural networks via proximal operators.
\newblock In {\em ICLR}, pages 1--19, 2019.

\bibitem{beltagy2020longformer}
I.~Beltagy, M.~E. Peters, and A.~Cohan.
\newblock Longformer: The long-document transformer.
\newblock {\em arXiv preprint arXiv:2004.05150}, 2020.

\bibitem{bengio2013estimating}
Y.~Bengio.
\newblock Estimating or propagating gradients through stochastic neurons.
\newblock {\em arXiv preprint arXiv:1305.2982}, 2013.

\bibitem{bulat2019xnor}
A.~Bulat and G.~Tzimiropoulos.
\newblock Xnor-net++: Improved binary neural networks.
\newblock In {\em BMVC}, page~62, 2019.

\bibitem{charikar2002similarity}
M.~S. Charikar.
\newblock Similarity estimation techniques from rounding algorithms.
\newblock In {\em STOC}, pages 380--388, 2002.

\bibitem{chen2020addernet}
H.~Chen, Y.~Wang, C.~Xu, B.~Shi, C.~Xu, Q.~Tian, and C.~Xu.
\newblock Addernet: Do we really need multiplications in deep learning?
\newblock In {\em CVPR}, pages 1468--1477, 2020.

\bibitem{choromanski2021rethinking}
K.~M. Choromanski, V.~Likhosherstov, D.~Dohan, X.~Song, A.~Gane, T.~Sarlos,
  P.~Hawkins, J.~Q. Davis, A.~Mohiuddin, L.~Kaiser, D.~B. Belanger, L.~J.
  Colwell, and A.~Weller.
\newblock Rethinking attention with performers.
\newblock In {\em ICLR}, pages 1--38, 2021.

\bibitem{twins}
X.~Chu, Z.~Tian, Y.~Wang, B.~Zhang, H.~Ren, X.~Wei, H.~Xia, and C.~Shen.
\newblock Twins: Revisiting the design of spatial attention in vision
  transformers.
\newblock In {\em NeurIPS}, pages 9355--9366, 2021.

\bibitem{courbariaux2015binaryconnect}
M.~Courbariaux, Y.~Bengio, and J.-P. David.
\newblock Binaryconnect: Training deep neural networks with binary weights
  during propagations.
\newblock In {\em NeurIPS}, volume~28, pages 3123--3131, 2015.

\bibitem{daras2020smyrf}
G.~Daras, N.~Kitaev, A.~Odena, and A.~G. Dimakis.
\newblock Smyrf-efficient attention using asymmetric clustering.
\newblock In {\em NeurIPS}, volume~33, pages 6476--6489, 2020.

\bibitem{datar2004locality}
M.~Datar, N.~Immorlica, P.~Indyk, and V.~S. Mirrokni.
\newblock Locality-sensitive hashing scheme based on p-stable distributions.
\newblock In {\em SoCG}, pages 253--262, 2004.

\bibitem{dehghani2018universal}
M.~Dehghani, S.~Gouws, O.~Vinyals, J.~Uszkoreit, and L.~Kaiser.
\newblock Universal transformers.
\newblock In {\em ICLR}, pages 1--23, 2019.

\bibitem{devlin2019bert}
J.~Devlin, M.~Chang, K.~Lee, and K.~Toutanova.
\newblock {BERT:} pre-training of deep bidirectional transformers for language
  understanding.
\newblock In J.~Burstein, C.~Doran, and T.~Solorio, editors, {\em NAACL-HLT},
  pages 4171--4186, 2019.

\bibitem{ding2019regularizing}
R.~Ding, T.-W. Chin, Z.~Liu, and D.~Marculescu.
\newblock Regularizing activation distribution for training binarized deep
  networks.
\newblock In {\em CVPR}, pages 11408--11417, 2019.

\bibitem{dosovitskiy2021an}
A.~Dosovitskiy, L.~Beyer, A.~Kolesnikov, D.~Weissenborn, X.~Zhai,
  T.~Unterthiner, M.~Dehghani, M.~Minderer, G.~Heigold, S.~Gelly, J.~Uszkoreit,
  and N.~Houlsby.
\newblock An image is worth 16x16 words: Transformers for image recognition at
  scale.
\newblock In {\em ICLR}, pages 1--21, 2021.

\bibitem{esser2020learned}
S.~K. Esser, J.~L. McKinstry, D.~Bablani, R.~Appuswamy, and D.~S. Modha.
\newblock Learned step size quantization.
\newblock In {\em ICLR}, pages 1--12, 2020.

\bibitem{gionis1999similarity}
A.~Gionis, P.~Indyk, and R.~Motwani.
\newblock Similarity search in high dimensions via hashing.
\newblock In {\em Vldb}, page 518–529, 1999.

\bibitem{gong2012iterative}
Y.~Gong, S.~Lazebnik, A.~Gordo, and F.~Perronnin.
\newblock Iterative quantization: A procrustean approach to learning binary
  codes for large-scale image retrieval.
\newblock {\em TPAMI}, 35(12):2916--2929, 2012.

\bibitem{han2016eie}
S.~Han, X.~Liu, H.~Mao, J.~Pu, A.~Pedram, M.~A. Horowitz, and W.~J. Dally.
\newblock Eie: Efficient inference engine on compressed deep neural network.
\newblock {\em ACM SIGARCH Computer Architecture News}, 44(3):243--254, 2016.

\bibitem{horowitz20141}
M.~Horowitz.
\newblock 1.1 computing's energy problem (and what we can do about it).
\newblock In {\em ISSCC}, pages 10--14, 2014.

\bibitem{hou2018loss}
L.~Hou and J.~T. Kwok.
\newblock Loss-aware weight quantization of deep networks.
\newblock In {\em ICLR}, pages 1--16, 2018.

\bibitem{hubara2016binarized}
I.~Hubara, M.~Courbariaux, D.~Soudry, R.~El-Yaniv, and Y.~Bengio.
\newblock Binarized neural networks.
\newblock In {\em NeurIPS}, pages 4107--4115, 2016.

\bibitem{hubara2017quantized}
I.~Hubara, M.~Courbariaux, D.~Soudry, R.~El-Yaniv, and Y.~Bengio.
\newblock Quantized neural networks: Training neural networks with low
  precision weights and activations.
\newblock {\em JMLR}, 18(1):6869--6898, 2017.

\bibitem{jaegle2021perceiver}
A.~Jaegle, F.~Gimeno, A.~Brock, O.~Vinyals, A.~Zisserman, and J.~Carreira.
\newblock Perceiver: General perception with iterative attention.
\newblock In {\em ICML}, pages 4651--4664, 2021.

\bibitem{jung2019learning}
S.~Jung, C.~Son, S.~Lee, J.~Son, J.-J. Han, Y.~Kwak, S.~J. Hwang, and C.~Choi.
\newblock Learning to quantize deep networks by optimizing quantization
  intervals with task loss.
\newblock In {\em CVPR}, pages 4350--4359, 2019.

\bibitem{katharopoulos2020transformers}
A.~Katharopoulos, A.~Vyas, N.~Pappas, and F.~Fleuret.
\newblock Transformers are rnns: Fast autoregressive transformers with linear
  attention.
\newblock In {\em ICML}, pages 5156--5165, 2020.

\bibitem{reformer}
N.~Kitaev, L.~Kaiser, and A.~Levskaya.
\newblock Reformer: The efficient transformer.
\newblock In {\em ICLR}, pages 1--12, 2020.

\bibitem{krizhevsky2012imagenet}
A.~Krizhevsky, I.~Sutskever, and G.~E. Hinton.
\newblock Imagenet classification with deep convolutional neural networks.
\newblock In {\em NeurIPS}, volume~25, pages 1106--1114, 2012.

\bibitem{kulis2011kernelized}
B.~Kulis and K.~Grauman.
\newblock Kernelized locality-sensitive hashing.
\newblock {\em TPAMI}, 34(6):1092--1104, 2011.

\bibitem{lian2019high}
X.~Lian, Z.~Liu, Z.~Song, J.~Dai, W.~Zhou, and X.~Ji.
\newblock High-performance fpga-based cnn accelerator with block-floating-point
  arithmetic.
\newblock {\em TVLSI}, 27(8):1874--1885, 2019.

\bibitem{lin2014fast}
G.~Lin, C.~Shen, Q.~Shi, A.~Van~den Hengel, and D.~Suter.
\newblock Fast supervised hashing with decision trees for high-dimensional
  data.
\newblock In {\em CVPR}, pages 1963--1970, 2014.

\bibitem{lin2017towards}
X.~Lin, C.~Zhao, and W.~Pan.
\newblock Towards accurate binary convolutional neural network.
\newblock In {\em NeurIPS}, pages 344--352, 2017.

\bibitem{liu2014discrete}
W.~Liu, C.~Mu, S.~Kumar, and S.-F. Chang.
\newblock Discrete graph hashing.
\newblock In {\em NeurIPS}, volume~27, pages 3419--3427, 2014.

\bibitem{liu2012supervised}
W.~Liu, J.~Wang, R.~Ji, Y.-G. Jiang, and S.-F. Chang.
\newblock Supervised hashing with kernels.
\newblock In {\em CVPR}, pages 2074--2081, 2012.

\bibitem{liu2011hashing}
W.~Liu, J.~Wang, S.~Kumar, and S.-F. Chang.
\newblock Hashing with graphs.
\newblock In {\em ICML}, pages 1--8, 2011.

\bibitem{liu2021swin}
Z.~Liu, Y.~Lin, Y.~Cao, H.~Hu, Y.~Wei, Z.~Zhang, S.~Lin, and B.~Guo.
\newblock Swin transformer: Hierarchical vision transformer using shifted
  windows.
\newblock In {\em ICCV}, pages 10012--10022, 2021.

\bibitem{Liu_2018_ECCV}
Z.~Liu, B.~Wu, W.~Luo, X.~Yang, W.~Liu, and K.-T. Cheng.
\newblock Bi-real net: Enhancing the performance of 1-bit cnns with improved
  representational capability and advanced training algorithm.
\newblock In {\em ECCV}, pages 722--737, 2018.

\bibitem{adamw}
I.~Loshchilov and F.~Hutter.
\newblock Decoupled weight decay regularization.
\newblock In {\em ICLR}, pages 1--18, 2019.

\bibitem{lu2021soft}
J.~Lu, J.~Yao, J.~Zhang, X.~Zhu, H.~Xu, W.~Gao, C.~Xu, T.~Xiang, and L.~Zhang.
\newblock Soft: Softmax-free transformer with linear complexity.
\newblock In {\em NeurIPS}, volume~34, pages 21297--21309, 2021.

\bibitem{martinez2020training}
B.~Martinez, J.~Yang, A.~Bulat, and G.~Tzimiropoulos.
\newblock Training binary neural networks with real-to-binary convolutions.
\newblock In {\em ICLR}, pages 1--11, 2020.

\bibitem{mishra2017apprentice}
A.~Mishra and D.~Marr.
\newblock Apprentice: Using knowledge distillation techniques to improve
  low-precision network accuracy.
\newblock In {\em ICLR}, pages 1--17, 2018.

\bibitem{norouzi2011minimal}
M.~Norouzi and D.~J. Fleet.
\newblock Minimal loss hashing for compact binary codes.
\newblock In {\em ICML}, pages 353--360, 2011.

\bibitem{peng2020random}
H.~Peng, N.~Pappas, D.~Yogatama, R.~Schwartz, N.~Smith, and L.~Kong.
\newblock Random feature attention.
\newblock In {\em ICLR}, pages 1--19, 2021.

\bibitem{qin2022bibert}
H.~Qin, Y.~Ding, M.~Zhang, Q.~YAN, A.~Liu, Q.~Dang, Z.~Liu, and X.~Liu.
\newblock Bi{BERT}: Accurate fully binarized {BERT}.
\newblock In {\em ICLR}, pages 1--24, 2022.

\bibitem{zhen2022cosformer}
Z.~Qin, W.~Sun, H.~Deng, D.~Li, Y.~Wei, B.~Lv, J.~Yan, L.~Kong, and Y.~Zhong.
\newblock cosformer: Rethinking softmax in attention.
\newblock In {\em ICLR}, pages 1--15, 2022.

\bibitem{NIPS2007_013a006f}
A.~Rahimi and B.~Recht.
\newblock Random features for large-scale kernel machines.
\newblock In {\em NeurIPS}, volume~20, pages 1177--1184, 2007.

\bibitem{rastegari2016xnor}
M.~Rastegari, V.~Ordonez, J.~Redmon, and A.~Farhadi.
\newblock Xnor-net: Imagenet classification using binary convolutional neural
  networks.
\newblock In {\em ECCV}, pages 525--542, 2016.

\bibitem{ren2021combiner}
H.~Ren, H.~Dai, Z.~Dai, M.~Yang, J.~Leskovec, D.~Schuurmans, and B.~Dai.
\newblock Combiner: Full attention transformer with sparse computation cost.
\newblock In {\em NeurIPS}, volume~34, pages 22470--22482, 2021.

\bibitem{sharma2018bit}
H.~Sharma, J.~Park, N.~Suda, L.~Lai, B.~Chau, V.~Chandra, and H.~Esmaeilzadeh.
\newblock Bit fusion: Bit-level dynamically composable architecture for
  accelerating deep neural network.
\newblock In {\em ISCA}, pages 764--775, 2018.

\bibitem{shen2020q}
S.~Shen, Z.~Dong, J.~Ye, L.~Ma, Z.~Yao, A.~Gholami, M.~W. Mahoney, and
  K.~Keutzer.
\newblock Q-bert: Hessian based ultra low precision quantization of bert.
\newblock In {\em AAAI}, volume~34, pages 8815--8821, 2020.

\bibitem{shu2021adder}
H.~Shu, J.~Wang, H.~Chen, L.~Li, Y.~Yang, and Y.~Wang.
\newblock Adder attention for vision transformer.
\newblock In {\em NeurIPS}, volume~34, pages 19899--19909, 2021.

\bibitem{sun2021sparse}
Z.~Sun, Y.~Yang, and S.~Yoo.
\newblock Sparse attention with learning to hash.
\newblock In {\em ICLR}, pages 1--20, 2022.

\bibitem{tay2020long}
Y.~Tay, M.~Dehghani, S.~Abnar, Y.~Shen, D.~Bahri, P.~Pham, J.~Rao, L.~Yang,
  S.~Ruder, and D.~Metzler.
\newblock Long range arena: A benchmark for efficient transformers.
\newblock In {\em ICLR}, pages 1--19, 2021.

\bibitem{touvron2021training}
H.~Touvron, M.~Cord, M.~Douze, F.~Massa, A.~Sablayrolles, and H.~J{\'e}gou.
\newblock Training data-efficient image transformers \& distillation through
  attention.
\newblock In {\em ICML}, pages 10347--10357, 2021.

\bibitem{vaswani2017attention}
A.~Vaswani, N.~Shazeer, N.~Parmar, J.~Uszkoreit, L.~Jones, A.~N. Gomez,
  {\L}.~Kaiser, and I.~Polosukhin.
\newblock Attention is all you need.
\newblock In {\em NeurIPS}, volume~30, pages 5998--6008, 2017.

\bibitem{vyas2020fast}
A.~Vyas, A.~Katharopoulos, and F.~Fleuret.
\newblock Fast transformers with clustered attention.
\newblock In {\em NeurIPS}, volume~33, pages 21665--21674, 2020.

\bibitem{wang2012semi}
J.~Wang, S.~Kumar, and S.-F. Chang.
\newblock Semi-supervised hashing for large-scale search.
\newblock {\em TPAMI}, 34(12):2393--2406, 2012.

\bibitem{linformer}
S.~Wang, B.~Li, M.~Khabsa, H.~Fang, and H.~Ma.
\newblock Linformer: Self-attention with linear complexity.
\newblock {\em arXiv preprint arXiv:2006.04768}, 2020.

\bibitem{wang2021pyramid}
W.~Wang, E.~Xie, X.~Li, D.-P. Fan, K.~Song, D.~Liang, T.~Lu, P.~Luo, and
  L.~Shao.
\newblock Pyramid vision transformer: A versatile backbone for dense prediction
  without convolutions.
\newblock In {\em ICCV}, pages 568--578, 2021.

\bibitem{pvtv2}
W.~Wang, E.~Xie, X.~Li, D.-P. Fan, K.~Song, D.~Liang, T.~Lu, P.~Luo, and
  L.~Shao.
\newblock Pvtv2: Improved baselines with pyramid vision transformer.
\newblock {\em Computational Visual Media}, 8(3):415--424, 2022.

\bibitem{weiss2008spectral}
Y.~Weiss, A.~Torralba, and R.~Fergus.
\newblock Spectral hashing.
\newblock In {\em NeurIPS}, volume~21, pages 1--8, 2008.

\bibitem{xiong2021nystromformer}
Y.~Xiong, Z.~Zeng, R.~Chakraborty, M.~Tan, G.~Fung, Y.~Li, and V.~Singh.
\newblock Nystr{\"o}mformer: A nyst{\"o}m-based algorithm for approximating
  self-attention.
\newblock In {\em AAAI}, volume~35, page 14138, 2021.

\bibitem{yan2020deep}
C.~Yan, B.~Gong, Y.~Wei, and Y.~Gao.
\newblock Deep multi-view enhancement hashing for image retrieval.
\newblock {\em TPAMI}, 43(4):1445--1451, 2020.

\bibitem{zadeh2020gobo}
A.~H. Zadeh, I.~Edo, O.~M. Awad, and A.~Moshovos.
\newblock Gobo: Quantizing attention-based nlp models for low latency and
  energy efficient inference.
\newblock In {\em MICRO}, pages 811--824, 2020.

\bibitem{zafrir2019q8bert}
O.~Zafrir, G.~Boudoukh, P.~Izsak, and M.~Wasserblat.
\newblock Q8bert: Quantized 8bit bert.
\newblock In {\em EMC2-NIPS}, pages 36--39, 2019.

\bibitem{zhou2016dorefa}
S.~Zhou, Y.~Wu, Z.~Ni, X.~Zhou, H.~Wen, and Y.~Zou.
\newblock Dorefa-net: Training low bitwidth convolutional neural networks with
  low bitwidth gradients.
\newblock {\em arXiv preprint arXiv:1606.06160}, 2016.

\bibitem{zhu1997algorithm}
C.~Zhu, R.~H. Byrd, P.~Lu, and J.~Nocedal.
\newblock Algorithm 778: L-bfgs-b: Fortran subroutines for large-scale
  bound-constrained optimization.
\newblock {\em ACM TOMS}, 23(4):550--560, 1997.

\bibitem{zhu2021long}
C.~Zhu, W.~Ping, C.~Xiao, M.~Shoeybi, T.~Goldstein, A.~Anandkumar, and
  B.~Catanzaro.
\newblock Long-short transformer: Efficient transformers for language and
  vision.
\newblock In {\em NeurIPS}, volume~34, pages 17723--17736, 2021.

\bibitem{zhuang2016fast}
B.~Zhuang, G.~Lin, C.~Shen, and I.~Reid.
\newblock Fast training of triplet-based deep binary embedding networks.
\newblock In {\em CVPR}, pages 5955--5964, 2016.

\bibitem{zhuang2019structured}
B.~Zhuang, C.~Shen, M.~Tan, L.~Liu, and I.~Reid.
\newblock Structured binary neural network for accurate image classification
  and semantic segmentation.
\newblock In {\em CVPR}, pages 413--422, 2019.

\end{thebibliography}
}

%%%%%%%%%%%%%%%%%%%%%%%%%%%%%%%%%%%%%%%%%%%%%%%%%%%%%%%%%%%%
\newpage
\section*{Checklist}

%%% BEGIN INSTRUCTIONS %%%
% The checklist follows the references.  Please
% read the checklist guidelines carefully for information on how to answer these
% questions.  For each question, change the default \answerTODO{} to \answerYes{},
% \answerNo{}, or \answerNA{}.  You are strongly encouraged to include a {\bf
% justification to your answer}, either by referencing the appropriate section of
% your paper or providing a brief inline description.  For example:
% \begin{itemize}
%   \item Did you include the license to the code and datasets? \answerYes{See Section~XXX.}
%   \item Did you include the license to the code and datasets? \answerNo{The code and the data are proprietary.}
%   \item Did you include the license to the code and datasets? \answerNA{}
% \end{itemize}
% Please do not modify the questions and only use the provided macros for your
% answers.  Note that the Checklist section does not count towards the page
% limit.  In your paper, please delete this instructions block and only keep the
% Checklist section heading above along with the questions/answers below.
%%% END INSTRUCTIONS %%%

\begin{enumerate}

\item For all authors...
\begin{enumerate}
  \item Do the main claims made in the abstract and introduction accurately reflect the paper's contributions and scope?
    \answerYes{See the abstract and introduction.}
  \item Did you describe the limitations of your work?
    \answerYes{See the conclusion in Section~\ref{sec:conclusion}.}
  \item Did you discuss any potential negative societal impacts of your work?
    \answerYes{See the conclusion in Section~\ref{sec:conclusion}.}
  \item Have you read the ethics review guidelines and ensured that your paper conforms to them?
    \answerYes{}
\end{enumerate}

\item If you are including theoretical results...
\begin{enumerate}
  \item Did you state the full set of assumptions of all theoretical results?
    \answerYes{}
        \item Did you include complete proofs of all theoretical results?
    \answerYes{}
\end{enumerate}

\item If you ran experiments...
\begin{enumerate}
  \item Did you include the code, data, and instructions needed to reproduce the main experimental results (either in the supplemental material or as a URL)?
    \answerYes{We have included the code and instructions needed in the supplemental material.}
  \item Did you specify all the training details (e.g., data splits, hyperparameters, how they were chosen)?
    \answerYes{See the implementation details in Section~\ref{sec:experiments}.}
        \item Did you report error bars (e.g., with respect to the random seed after running experiments multiple times)?
    \answerNo{Error bars are not reported because it would be too computationally expensive. We use the same random seed as in recent works for fair comparison.}
        \item Did you include the total amount of compute and the type of resources used (e.g., type of GPUs, internal cluster, or cloud provider)?
    \answerYes{See the implementation details in Section~\ref{sec:experiments}.}
\end{enumerate}

\item If you are using existing assets (e.g., code, data, models) or curating/releasing new assets...
\begin{enumerate}
  \item If your work uses existing assets, did you cite the creators?
    \answerYes{See Section~\ref{sec:experiments}.}
  \item Did you mention the license of the assets?
     \answerNA{}
  \item Did you include any new assets either in the supplemental material or as a URL?
    \answerYes{Code is included in the supplementary material.}
  \item Did you discuss whether and how consent was obtained from people whose data you're using/curating?
    \answerNo{We use public datasets.}
  \item Did you discuss whether the data you are using/curating contains personally identifiable information or offensive content?
    \answerNA{}
\end{enumerate}

\item If you used crowdsourcing or conducted research with human subjects...
\begin{enumerate}
  \item Did you include the full text of instructions given to participants and screenshots, if applicable?
    \answerNA{}
  \item Did you describe any potential participant risks, with links to Institutional Review Board (IRB) approvals, if applicable?
    \answerNA{}
  \item Did you include the estimated hourly wage paid to participants and the total amount spent on participant compensation?
    \answerNA{}
\end{enumerate}

\end{enumerate}

%%%%%%%%%%%%%%%%%%%%%%%%%%%%%%%%%%%%%%%%%%%%%%%%%%%%%%%%%%%%

% \appendix
% \newpage
% \input{supp}

\end{document}

\begin{center}
	{
		\Large{\textbf{Appendix}}
	}
\end{center}

\renewcommand\thesection{\Alph{section}}
\renewcommand\thefigure{\Alph{figure}}
\renewcommand\thetable{\Alph{table}}
\renewcommand{\theequation}{\Alph{equation}}

\setcounter{table}{0}
\setcounter{figure}{0}

We organize our supplementary material as follows. 
\begin{itemize}[leftmargin=*]
    % \item In Section~\ref{sec:complexity}, we include more details on how to compute the computational complexity.
    % \item In Section~\ref{sec:more_results_imagenet}, we provide more results on ImageNet-1K.
    \item In Section~\ref{sec:cost_saving_attention}, we show the complexity and energy footprint saving for the attention layers only.
    \item In Section~\ref{sec:throughput_imagenet}, we provide more throughput results on ImageNet-1K.
    % \item In Section~\ref{sec:explore_training_from_scratch}, we explore the effect of the training from scratch scheme.
    % \item In Section~\ref{sec:effect_m}, we study the effect of different $m$.
\end{itemize}

\section{Cost Saving for Attentions Only}
\label{sec:cost_saving_attention}
In this section, we show the results of cost saving for attentions only using different architectures on ImageNet-1K. From Table~\ref{tab:cost_saving_attention}, our EcoFormer consistently saves massive \revise{on-chip} energy footprint. In particular, on PVTv2-B0, we save 93\% \revise{on-chip} energy footprint. On larger models, the saving is still significant. Note that larger models 
also come with more computational cost from  
the linear projection layers in MSAs, as shown in Figure~2 in the main paper.
Since our EcoFormer does not target these projection layers, larger models save slightly less energy in MSAs compared to those of smaller models.

\begin{table}[htb!]
\centering
\renewcommand\arraystretch{1.3}
\caption{The cost saving for attentions only, excluding other types of layers. The number of multiplications, additions, as well as \revise{on-chip} energy consumption are calculated based on an image of resolution $224\times224$.}
\label{tab:cost_saving_attention}
\scalebox{0.9}{
\begin{tabular}{l|c|ccll}
Model   & Method & \#Mul. (B) & \#Add. (B) & Energy (B pJ) \\ \shline
\multirow{2}{*}{PVTv2-B0~\cite{pvtv2}}   & MSA    & 1.58     & 1.56     & 7.26              \\
 & \textbf{Ours} &  \textbf{0.10}    & \textbf{0.13}     &  \textbf{0.50} \textcolor{blue}{(\textbf{-93\%})}          \\   \hline
 \multirow{2}{*}{PVTv2-B1}    & MSA    & 3.38     & 3.36     & 15.52     \\
 & \textbf{Ours} & \textbf{0.39}     & \textbf{0.45}     & \textbf{1.84} \textcolor{blue}{(\textbf{-88\%})}         \\   \hline
 \multirow{2}{*}{PVTv2-B2}    & MSA    &    5.59  &  5.56    &   25.69  \\
 & \textbf{Ours} & \textbf{0.80}     &  \textbf{0.92}    &  \textbf{3.80} \textcolor{blue}{(\textbf{-85\%})}        \\   \hline
\multirow{2}{*}{PVTv2-B3}    & MSA    & 6.85     &  6.81    & 31.49  \\
 & \textbf{Ours} & \textbf{1.53}     & \textbf{1.71}     & \textbf{7.21} \textcolor{blue}{(\textbf{-77\%})} \\   \hline
\multirow{2}{*}{PVTv2-B4}    & MSA    & 8.63    & 8.59     & 39.69 \\ 
 & \textbf{Ours} & \textbf{2.24}    & \textbf{2.49}     & \textbf{10.52} \textcolor{blue}{(\textbf{-74\%})}  \\ \hline
 \multirow{2}{*}{Twins-SVT-S~\cite{twins}}    & MSA    & 4.01     & 3.96     & 18.41\\
 & \textbf{Ours} &  \textbf{0.77}    & \textbf{0.86}     &  \textbf{3.63} \textcolor{blue}{(\textbf{-80\%})} \\   \hline
\end{tabular}
}
\vspace{-2mm}
\end{table} 

\section{More Throughput Results on ImageNet-1K}
\label{sec:throughput_imagenet}
To show the actual inference speed on a hardware device, we measure the throughput of different methods on a single NVIDIA RTX 3090 GPU. We compare EcoFormer with the standard multi-head self-attention (MSA) and kernel-based linear attention (KLA)~\cite{choromanski2021rethinking}. 
From Table~\ref{tab:throughput}, KLA shows higher throughput than MSA, while our EcoFormer achieves even faster throughput than KLA, thanks to the reduced feature dimensions ($b$ vs. $D_p$) of queries and keys.
With efficient energy-efficient accumulation implementation, the throughput of our EcoFormer can be further improved, which will be explored in the future.

\begin{table}[htb!]
\centering
\renewcommand\arraystretch{1.3}
\caption{Throughput (images/s) of different methods on ImageNet-1K. MSA denotes the standard multi-head self-attention and KLA represents the kernel-based linear attention. The throughput is measured with a mini-batch size of 32 and an image resolution of $224 \times 224$ on a single NVIDIA RTX 3090 GPU.}
\label{tab:throughput}
\scalebox{0.9}{
\begin{tabular}{c|ccccc}
% Model   & Method & Test Throughput (images/s) \\ \shline
% \multirow{3}{*}{PVTv2-B0} & MSA & 850 \\ 
% & KLA~\cite{choromanski2021rethinking} & 1166 \\
% & \textbf{Ours} & \textbf{1379} \\
% \hline
% \multirow{3}{*}{PVTv2-B1} & MSA & 621 \\ 
% & KLA & 769 \\
% & \textbf{Ours} & \textbf{874} \\
% \hline
% \multirow{3}{*}{PVTv2-B2} & MSA & 404 \\ 
% & KLA & 444 \\
% & \textbf{Ours} & \textbf{483} \\
% \hline
% \multirow{3}{*}{Twins-SVT-S} & MSA & 426 \\ 
% & KLA & 489 \\
% & \textbf{Ours} & \textbf{576} \\
Method & PVTv2-B0 & PVTv2-B1 & PVTv2-B2 & Twins-SVT-S \\ \shline
MSA & 850 & 621 & 404 & 426 \\
KLA~\cite{choromanski2021rethinking} & 1166 & 769 & 444 & 489 \\
Ours & \textbf{1379} & \textbf{874} & \textbf{483} & \textbf{576} \\
\end{tabular}
}
\vspace{-2mm}
\end{table}

\bibliographystyle{abbrv}
{
    \small
	\bibliography{reference}
}